\documentclass[conference]{IEEEtran}
\IEEEoverridecommandlockouts

\usepackage{cite}
\usepackage{amsmath,amssymb,amsfonts}
\usepackage{algorithmic}
\usepackage{graphicx}
\usepackage{textcomp}
\usepackage{xcolor}
\usepackage{subfiles}
\usepackage{booktabs}
\usepackage{float}
\usepackage{comment}
\usepackage{hyperref}
\def\BibTeX{{\rm B\kern-.05em{\sc i\kern-.025em b}\kern-.08em
    T\kern-.1667em\lower.7ex\hbox{E}\kern-.125emX}}

\begin{document}

\title{Task-Driven Semantic Quantization and Imitation Learning for Goal-Oriented Communications}

\author{\IEEEauthorblockN{Yu-Chieh Chao$^1$, Yubei Chen$^1$, Weiwei Wang$^1$, Achintha Wijesinghe$^1$, Suchinthaka Wanninayaka$^1$, \\
 Songyang Zhang$^2$, \textit{Member, IEEE}, and Zhi Ding$^1$, \textit{Fellow,~IEEE}}
\IEEEauthorblockA{$^1$University of California at Davis, Davis, CA, USA \\ $^2$University of Louisiana at Lafayette, Lafayette, LA, USA}\\
\thanks{This material is based upon work supported by the National Science Foundation under Grant No. 2332760 and No. 2349878.}\\
\vspace*{-20mm}}

\maketitle


\begin{abstract}
Semantic communication marks a new paradigm shift from 
bit-wise data transmission to semantic 
information delivery for the purpose of bandwidth reduction. To more effectively carry out specialized downstream tasks at the receiver end, it is crucial to define the most critical semantic message in the data based on the task or goal-oriented features.
In this work, we propose a novel goal-oriented communication (GO-COM) framework, namely Goal-Oriented Semantic Variational Autoencoder (GOS-VAE), by focusing on the extraction of the semantics vital to the downstream tasks. Specifically, 
we adopt a Vector Quantized Variational Autoencoder (VQ-VAE) to compress media data at the transmitter side. Instead of targeting the pixel-wise image data reconstruction, we measure the quality-of-service at the receiver end based on
a pre-defined task-incentivized model. Moreover, to capture the relevant semantic features in the data reconstruction, imitation learning is adopted to measure the data regeneration quality in terms of
goal-oriented semantics. Our experimental results 
demonstrate the power of imitation learning in characterizing goal-oriented semantics and bandwidth efficiency of our proposed GOS-VAE.

\end{abstract}


\vspace*{-1mm}
\section{Introduction}
\vspace*{-1mm}


Next-generation wireless networks and Artificial Intelligence (AI) algorithms have found 
a wide range of data-intensive applications, 
including augmented and virtual reality \cite{hazarika2023towards}, and autonomous driving \cite{chen2020vision}, where low-latency data transport
and high-accuracy decision-making play crucial roles. As these applications proliferate, the hunger for bandwidth and
data rates continues to grow, straining the already 
scarce communication resources \cite{strinati20216g}. To achieve bandwidth efficiency and ensure data transmission quality, the concept of semantic communications has recently 
re-surfaced as an important paradigm, which aims at conveying the most critical semantic information rather than bit-wise packet transport \cite{luo2022semantic}.

Semantic communication systems commonly employ deep learning techniques for embedding representations and regenerating data. For example, DeepSC \cite{xie2021deep} focused on communicating the semantic meaning in text messages by utilizing a
deep-learning transformer architecture. 
Beyond text communications, 
authors of \cite{bourtsoulatze2019deep} proposed a joint source and channel coding technique for wireless image transmission, using a Convolutional Neural Network (CNN) to directly map image pixels to channel input symbols. Another work (VQ-DeepSC \cite{fu2023vector})
presented a vector quantization semantic 
communication system to compress 
multi-scale semantic features through codebook quantization. 
The recent development of generative AI has further
inspired other generative frameworks for 
semantic communications. For example, in \cite{lokumarambage2023wireless}, a pre-trained Generative Adversarial Network (GAN) is utilized to reconstruct images at the receiver. Similarly, Generative Semantic Communication (GESCO) introduced in \cite{grassucci2023generative} utilizes a diffusion-based architecture for data generation based on segmentation maps.
Nevertheless, classic semantic communication frameworks continue to stress the visual quality of the reconstructed images. Future communication systems are expected to play an increasingly important part to serve automation, artificial intelligence, and other decision-making applications, instead of being a pipe to provide data to only human end-users. Thus, next-generation networks must prioritize task-driven semantic communication for downstream tasks without involving human viewers, making efficient semantic feature extraction crucial.

\begin{figure*}[t]
\centering
\includegraphics[width=0.9\textwidth]{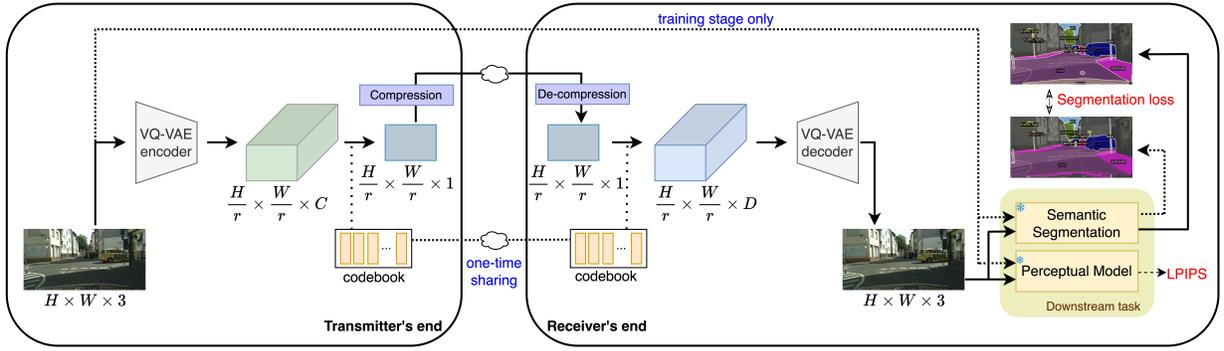}
\caption{The proposed Goal-Oriented Semantic Variational Autoencoder (GOS-VAE) framework for Semantic Communication: The snowflakes symbol denotes pre-trained and fixed model parameters.}\vspace*{-3mm}
\label{fig:sqvae}\vspace*{-3mm}
\end{figure*}

Recognizing the key role of communication networks in AI-driven data applications, recent works have shifted towards task/goal-oriented semantic communication (GO-COM) systems \cite{kang2022task}. Unlike classic semantic communications, these GO-COMs focus more on the delivery of key semantic information for the specific downstream tasks at the receiver end.

For this, \cite{wu2024semantic} proposed a semantic image transmission system that allocates higher data rates to Regions of Interest (ROI) within images. However, ROIs can vary significantly across downstream tasks. The TasCom framework \cite{fu2024generative} addresses this by transmitting only task-specific features, using an Adaptive Coding Controller to prioritize those most relevant to AI performance. Similarly, VIS-SemCom \cite{lv2024importance} reduces redundancy by transmitting feature maps of key objects, such as vehicles and pedestrians, in scenes like autonomous driving. Despite these progresses, existing GO-COMs reconstruct data using pre-defined semantic features, which lack adaptivity and generalization. For instance, while trees may generally be considered non-essential in autonomous driving, an accidentally fallen tree branch obstructing the road would require immediate attention. Furthermore, onboard resources are often insufficient to manage multiple complex tasks with models containing billions of parameters \cite{achiam2023gpt, touvron2023llama, liu2024visual}. Thus, critical challenges remain in advancing GO-COMs, particularly in achieving efficient semantic extraction and implementation.

To address the aforementioned challenges, this work introduces a novel GO-COM framework, namely Goal-Oriented Semantic Variational Autoencoder (GOS-VAE), integrating the Vector Quantized VAE (VQ-VAE) and imitation learning. 
We summarize our major contributions 
as follows:

\begin{itemize}
  \item Taking autonomous driving as an exemplary application, we propose the novel GOS-VAE framework supported
  by a VQ-VAE backbone, which defines the goal-oriented ``semantics" beneficial to the downstream task.
  \item To save computation, GOS-VAE places
  the decoder and computation-intensive downstream task model on a powerful back-end server while deploying a low-complexity encoder at the sender to improve efficiency.
  \item To preserve information vital
  to downstream tasks, our GOS-VAE
adopts imitation learning capable of processing different tasks
without manual labeling.
  \item Through flexible adjustment of network depth and codebook size, plus 
  a customized shallow CNN structure, our proposed GOS-VAE achieves superior
 signal recovery performance at reduced bandwidth
  consumption.
\end{itemize}


\vspace*{-1mm}
\section{Method}
\vspace*{-1mm}

We begin by introducing the structure of GOS-VAE \footnote{The code is available at \url{https://github.com/JayChao0331/GOS-VAE.git}}.

\subsubsection{Objective}
Our GOS-VAE focuses on application scenarios like autonomous driving, where edge devices such as autonomous vehicles or robots host limited computational power and is connected
to back-end servers with powerful computing capacities. Particularly, we use image transmission for remote vehicle control as an exemplary application. Though demonstrated on image transmission, our framework extends to AI-driven platforms, with learning-based source embedding adaptable to diverse channel models and protocols.

\subsubsection{Overall Structure}
The overall structure of our GOS-VAE is illustrated as Fig.~\ref{fig:sqvae}.
The transmitter features a lightweight encoder to embed the original images before transmitting a compressed representation to the receiver. The receiver reconstructs images for the accomplishment of its downstream tasks. Inspired by \cite{fu2023vector}, a Vector Quantized-Variational AutoEncoder (VQ-VAE) \cite{van2017neural} structure forms the backbone of the encoder-decoder model. Instead of focusing only on the visual quality of reconstructed images, we train a codebook to characterize
semantic information for the downstream task. During training, we end-to-end optimize the entire framework through imitation learning with performance feedback from the downstream task. 
This work focuses on a single task—semantic segmentation—leaving multi-task generalizations for future work. In the following sections, we introduce each functionality of our proposed GOS-VAE.

\vspace*{-2mm}
\subsection{VQ-VAE Backbone}
Our GOS-VAE uses VQ-VAE as encoder and decoder, deployed at the transmitter and the receiver, respectively.
The transmitter input is an image $x \in \mathbb{R}^{H \times W \times 3}$, where $H$, $W$, and $3$ denote the height, width, and the three RGB channels, respectively. The VQ-VAE encoder $z_{e}$ compresses the image into a feature map $z_{e}(x) \in \mathbb{R}^{\frac{H}{r} \times \frac{W}{r} \times C}$, where $r$ is the compression ratio affected by the multi-layer Convolutional Neural Network (CNN). Given the feature map, we use a codebook to further reduce transmission data payload. The channel dimension is reduced by a factor of $C$. Given a codebook $\textbf{e}=\{e_i\}_{i=1}^{K} \in \mathbb{R}^{K \times D}$ of size $K$ and codeword length $D$, a quantized map $z \in \mathbb{R}^{\frac{H}{r} \times \frac{W}{r}}$ based on a nearest-neighbor lookup can be found:
\begin{equation}
q(z_{ij} = k|x) = 
\begin{cases} 
1 & \text{for } k = \arg \min_{\ell} \| z_e(x) - e_{\ell} \|_2 , \\
0 & \text{otherwise}
\end{cases}.
\end{equation}

Both the transmitter and the receiver should
pre-store the learned codebook before real-time networking.
After quantization, the transmitter sends the codeword index to the receiver which converts the
codeword $z$ back to its feature map, denoted as $z_{q}(x) \in \mathbb{R}^{\frac{H}{r} \times \frac{W}{r} \times D}$, which is calculated as
\begin{equation}
z_q(x)_{ij} = e_k, \quad \text{where} \quad k = \arg \min_{\ell} \| z_e(x) - e_{\ell} \|_2 .
\end{equation}
The receiver forwards the feature map to its CNN decoder to recover the spatial dimension of the feature map and finally reconstruct the image $\hat{x} \in \mathbb{R}^{H \times W \times 3}$.

\vspace*{-2mm}
\subsection{Post-Training Communication Showtime}
During showtime, the transmitter only sends the codeword map $z$ for each captured RGB image $x$ to the receiver under compression ratio $r$. This compressed representation reduces bandwidth usage while preserving essential information. {Furthermore, due to the skewed distribution of the learned codebook, the codeword map may be further compressed by entropy encoding.} Our proposed framework
trains the compression network VQ-VAE in an end-to-end manner in view of the downstream task to ensure that 
the codeword map retains both good perceptual quality and the semantic information vital to
downstream task performance. The details will be introduced in Section \ref{sec:obj}.

\vspace*{-2mm}
\subsection{Downstream Task}
In this work, we use semantic segmentation as downstream task, which identifies the object category of each pixel. For this task, we adopt an open-source and pre-trained OneFormer \cite{jain2023oneformer} that unifies semantic, instance, and panoptic segmentation within a single model. Given an image $x \in \mathbb{R}^{H \times W \times 3}$, the pre-trained OneFormer $F$ predicts the segmentation map as
\begin{equation}
\vspace*{-2mm}
S = \mbox{Softmax}(F(x)) \in \mathbb{R}^{H \times W \times m},
\end{equation}
where $m$ is the number of object classes. The pre-trained OneFormer $F$ is unchanged during the training phase. To leverage the information from the downstream tasks, 
reconstructed images are fed into the pre-trained networks to generate additional loss terms for deep training. Additional details are explained in the following sections.

\vspace*{-2mm}

\subsection{Imitation Learning} \label{sec:il}
Not relying on ground-truth segmentation labels, we apply imitation learning \cite{xu2023bits} to guide the learning process. The goal of GOS-VAE is to reconstruct an image semantically similar to the original. For example, if the original image’s segmentation map identifies a vehicle resembling an ambulance, GOS-VAE should reconstruct a visually similar vehicle. This ensures the preservation of critical semantic features, such as vehicle characteristics. Imitation learning could also eliminate the need for ground-truth labeling and manual intervention.

In our framework, the original image $x$ first passes through the segmentation model $F$ to generate the segmentation map $S \in \mathbb{R}^{H \times W \times m}$, which serves as the imitation target. The reconstruction from GOS-VAE $\hat{x}$ is also given to the segmentation model $F$ to define its segmentation map $\hat{S} \in \mathbb{R}^{H \times W \times m}$. The learning objective is to minimize the difference of distributions between the two segmentation maps.

\vspace*{-2mm}

\subsection{Objective Function} \label{sec:obj}

For the VQ-VAE, the original loss function is designed for pixel-wise reconstruction, i.e.,
for $N = H \times W \times 3$,
\begin{equation}
L_{v} = \frac{1}{N} \sum_{i=1}^{N} (x_i - \hat{x}_i)^2 + \|\text{sg}[z_e(x)] - e\|_2^2 + \beta \|z_e(x) - \text{sg}[e]\|_2^2 ,
\end{equation}
in which $\beta=0.25$ is a constant, and `$\text{sg}$' denotes the stop-gradient operator.
The first term captures the Mean-Squared Error (MSE) between the two images and approximates the reconstruction loss $\log p(x|z_q(x))$. Since the gradient does not flow through the encoder due to the quantization step, the second and third terms of the loss are used to optimize the codebook and the encoder, respectively.

Since the proposed GOS-VAE focuses on the efficacy of downstream tasks 
according to the reconstruction, we
replace the pixel-wise MSE with a task-incentivized loss. Specifically, given two predicted segmentation maps, where each pixel represents a probability distribution over object categories, we compute the distribution distance using Jensen-Shannon Divergence (JSD), as described in Section \ref{sec:il}.
Since OneFormer is a pre-trained and fixed segmentation model, its data distribution aligns with the training datasets. Without further perceptual constraint on the reconstruction, GOS-VAE would struggle to perform well on the downstream task. 
To address this issue, we incorporate the Learned Perceptual Image Patch Similarity (LPIPS) \cite{zhang2018unreasonable} as a perceptual regularization in the loss function, using a pre-trained VGG16 \cite{simonyan2014very}. LPIPS compares two images by measuring the differences of their feature maps from the pre-trained VGG16. This term captures perceptual similarity, focusing on high-level semantic differences instead of pixel-wise changes. Finally, our objective function for the proposed GOS-VAE is
\begin{align}
L_{s} = & \, \text{LPIPS}(x, \hat{x}) + D_{JS}(S \parallel \hat{S}) \nonumber \\
        & + \|\text{sg}[z_e(x)] - e\|_2^2 + \beta \|z_e(x) - \text{sg}[e]\|_2^2 ,
\label{eq:5}
\end{align}
where $D_{JS}(S \parallel \hat{S}) = \frac{1}{2} D_\text{KL}(S \parallel M) + \frac{1}{2} D_\text{KL}(\hat{S} \parallel M)$, $M = \frac{1}{2}(S + \hat{S})$, and $D_{KL}$ denotes 
the Kullback-Leibler Divergence.

In this work, we provide several alternative training schemes for the proposed GOS-VAE. In the basic GOS-VAE, we train the model
from scratch using the objective function $L_{s}$. 
For an upgrade model GOS-VAE$^{*}$, we first pre-train a VQ-VAE using the original objective function $L_{v}$ before fine-tuning based on the objective function $L{s}$.
We further tested new models, VQ-VAE$^{\dagger}$ and the corresponding GOS-VAE$^{\dagger}$,  by replacing the CNN layers with Residual Blocks \cite{he2016deep} and by increasing
the codebook size. 
All these designs are studied, together with comprehensive discussions in Section \ref{sec:exp}.


\setlength{\tabcolsep}{6pt}

\section{Experiments \& Analyses}
\subsection{Experimental Settings} \label{sec:exp}
\vspace*{-1mm}
\subsubsection{Dataset} We test GOS-VAE using two datasets: Cityscapes \cite{cordts2016cityscapes} and ADE20K \cite{zhou2019semantic}. Cityscapes contains $2,975$ training and $500$ validation images, featuring high-resolution urban street scenes, annotated with $35$ object categories for semantic segmentation. Among these object categories, $19$ are considered
after the pre-training setup of OneFormer \cite{jain2023oneformer}. ADE20K includes $20,100$ training and $2,000$ validation images with diverse 
indoor and outdoor scenes, covering $150$ object categories for segmentation. For training, images are resized to $256 \times 512$, with horizontal flipping augmentation.

\subsubsection{Baselines}
We compare with conventional JPEG, Autoencoder, VQ-VAE \cite{van2017neural}, VQ-GAN \cite{esser2021taming}, and diffusion-based GESCO \cite{grassucci2023generative} using semantic segmentation as the downstream task. We also evaluate alternative GOS-VAE training schemes, including a basic version with shallow CNN layers, GOS-VAE$^{*}$, and GOS-VAE$^{\dagger}$. Specifically, in GOS-VAE$^{\dagger}$, we replaced the shallow CNNs with Residual Blocks \cite{he2016deep} and increased the codebook size from $512$ to $12,000$, to test the performance upper bound of Imitation Learning.

\vspace*{-2mm}
\begin{table}[htb]
\centering
\vspace{-1mm}
\caption{Performance comparisons on the Cityscapes semantic segmentation dataset.}
\vspace{-2mm}
\begin{tabular}{lcccc}
\toprule
\textbf{Models}                          & \textbf{Bandwidth} & \textbf{\# params} & \textbf{mIoU \textuparrow} & \textbf{Accuracy \textuparrow} \\
  &  \textbf{(KB)}  &  \textbf{(M)}  &  \textbf{(\%)}  &  \textbf{(\%)}  \\
\midrule
JPEG &  11.469  &  -  &  40.134  &  84.706 \\
Autoencoder  &  12.833  &  0.13  &  12.924  &  48.418  \\
VQ-GAN (r=4)     & 6.791     &  24.00                    & 54.238 & 92.523     \\
GESCO                      & 14.526     &  674.71                    & {\underline{58.969}}               & \textbf{95.351}                   
              \\
VQ-VAE (r=4)                      
      & 7.447  &  0.70             & 53.040                            & 91.647                \\
VQ-VAE$^\dagger$ (r=4)        & 8.309  &  7.32       & 54.961                            & 92.762                \\
GOS-VAE (r=4)  &  8.321  &  0.70 &  57.342  &  93.176  \\
GOS-VAE$^{*}$ (r=4)      & 8.385  &  0.70                   & 57.612            & 93.209                      \\
GOS-VAE$^\dagger$ (r=4)                &
10.092    &  7.32   & \textbf{61.318}
      & {\underline{94.087}}         \\

\bottomrule
\\
\end{tabular}\vspace*{-4mm}
\label{tab:table1}
\end{table}

\vspace*{+2mm}
\subsubsection{Setups}
For the Cityscapes dataset, both VQ-VAE and GOS-VAE are trained for $500$ epochs. GOS-VAE$^{*}$ and GOS-VAE$^{\dagger}$ are initialized using the pre-trained VQ-VAE and VQ-VAE$^{\dagger}$, respectively, before
fine-tuning for $100$ more epochs. For both JPEG and Autoencoder, we adjust the compression ratio to obtain comparable transmission bandwidth or payload. The CNN-based Autoencoder is trained for $500$ epochs. The VQ-GAN is trained for $272$ epochs. GESCO is trained for $250,000$ steps, equivalent to approximately $125$ epochs, with a diffusion step of $100$. Note that all $35$ segmentation categories are utilized for GESCO training and testing to reproduce its performance without changing the proposed architecture. 

For the ADE20K dataset, VQ-VAE and GOS-VAE are trained for $100$ epochs, while GOS-VAE$^{*}$ and GOS-VAE$^{\dagger}$ are initialized using the pre-trained VQ-VAE and VQ-VAE$^{\dagger}$ weights and fine-tuned for an additional $40$ epochs. The CNN-based Autoencoder is trained for $100$ epochs. The VQ-GAN is trained for $32$ epochs. GESCO is trained for $300,000$ steps, equivalent to approximately $15$ epochs, with a diffusion step of $100$. Each of the proposed GOS-VAE alternatives is trained and tested using a single RTX 4090 GPU, while GESCO is trained on an A100 GPU.

\begin{table}[t]
\centering
\caption{Performance comparisons on the ADE20K semantic segmentation dataset.}
\vspace{-2mm}
\begin{tabular}{lccc}
\toprule
\textbf{Models}   &  \textbf{\# params (M)}                        & \textbf{mIoU (\%) \textuparrow} & \textbf{Accuracy (\%) \textuparrow} \\
\midrule
JPEG  &  -  &  32.111  &  77.474  \\
Autoencoder  &  0.13  &  22.389  &  69.607  \\
VQ-GAN (r=4)   &  24.00    &    38.738      &  80.674                      \\
GESCO   &  681.33                                 & 16.850                   & 61.170                      \\
VQ-VAE (r=4)   &  0.70                                & 38.366                   & 80.333                      \\
VQ-VAE$^{\dagger}$ (r=4)  &  7.32  &  38.612  &  80.887  \\
GOS-VAE (r=4)  &  0.70  &  39.075  &  80.974  \\
GOS-VAE$^{*}$ (r=4)  &  0.70  &              \underline{39.128}    & \underline{81.019}                  \\
GOS-VAE $^{\dagger}$ (r=4)  &  7.32  &  \textbf{40.765}  &  \textbf{81.974}  \\
\bottomrule
\end{tabular}
\label{tab:table2} \vspace*{-4mm}
\end{table}

\begin{figure*}[t]
\centering
\includegraphics[width=\textwidth,height=0.2\textwidth
]{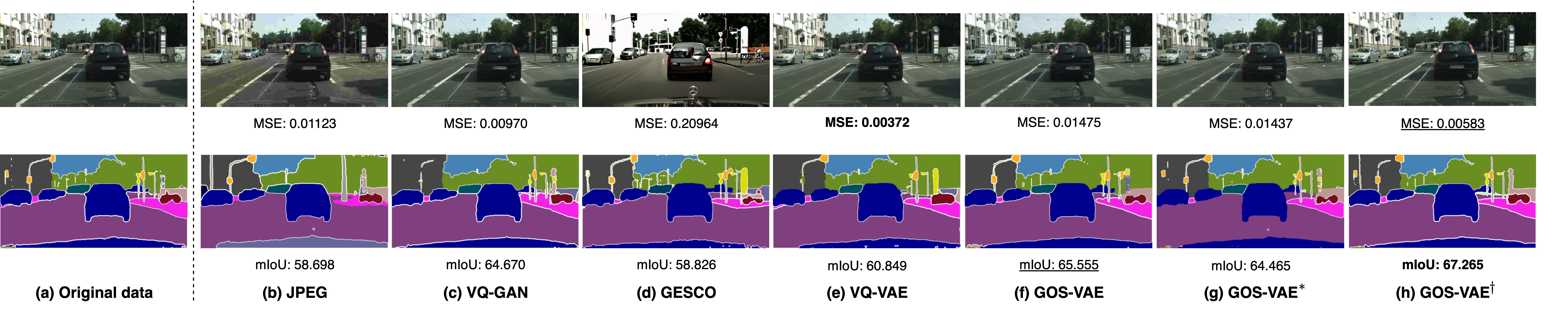}
\vspace*{-4mm}
\caption{Visualization Results of Different Methods on Cityscapes Dataset.}
\label{fig:visualization}
\vspace*{-4mm}
\end{figure*}

\vspace{-1mm}
\subsection{Overall Reconstruction Quality}
\vspace{-1mm}
We first compare the overall quality of image reconstruction of different GO-COM frameworks. 
Particularly, we use the performance of downstream image segmentation as evaluation metrics, in
terms of mean Intersection over Union (mIoU) and pixel-wise accuracy. We measure the bandwidth (payload) of the compressed data required by the receiver for image reconstruction. 
We compare the three different versions of our GOS-VAE against the Autoencoder, VQ-VAE, VQ-GAN, diffusion-based GESCO, together
with the classic JPEG compression.

The experimental results 
are presented in Table \ref{tab:table1}. Although performance can be easily improved by using more bandwidth at a compression ratio of $r=2$, we focus on $r=4$ here to prioritize bandwidth efficiency. 
More specifically, we set all the methods with a similar payload for better illustration.
From Table \ref{tab:table1}, we first notice that VQ-VAE surpasses JPEG and Autoencoder in both improving performance and bandwidth reduction,
owing to the use of codebook quantization. Moreover, we also notice that VQ-GAN achieves similar performance to VQ-VAE with more realistic image generation due to the discriminator. However, this addition also results in a significantly larger model size and longer training time. Next, by comparing GOS-VAE to VQ-VAE and VQ-GAN, we showcase the benefit of end-to-end learning in the compression network alongside the downstream task, as it preserves the semantic information critical to downstream task performance. Compared to the diffusion-based GESCO, we further observe that GOS-VAE achieves comparable performance while consuming significantly less bandwidth at a compression ratio $r=4$. Note that GESCO requires the ground-truth semantic segmentation map and edge map 
to generate the corresponding image during both training and testing phases. Moreover, GESCO utilizes semantic segmentation maps with $35$ object categories, utilizing more detailed information for image generation. In contrast, our method only uses $19$ object categories, following the pre-trained OneFormer setup \cite{jain2023oneformer}.

\begin{table}[htb]
\centering
\vspace*{-3mm}
\caption{Quantitative Comparisons of Visual Quality in Reconstructed Images using the Cityscapes dataset.}
\small
\begin{tabular}{l@{\hskip 4pt}c@{\hskip 4pt}c@{\hskip 4pt}c@{\hskip 8pt}c@{\hskip 4pt}c@{\hskip 4pt}c}
\toprule
 & \multicolumn{3}{c}{\textbf{Cityscapes}} & \multicolumn{3}{c}{\textbf{ADE20K}} \\
\cmidrule(lr){2-4} \cmidrule(lr){5-7}
\textbf{Models} & \textbf{MSE $\downarrow$} & \textbf{FID $\downarrow$} & \textbf{LPIPS $\downarrow$} & \textbf{MSE $\downarrow$} & \textbf{FID $\downarrow$} & \textbf{LPIPS $\downarrow$} \\
\midrule
VQ-GAN & 0.006 & \underline{20.212} & \underline{0.130} & 0.012 & \underline{12.383} & 0.177 \\
GESCO & 0.355 & 71.893 & 0.544 & 0.549 & 92.317 & 0.725 \\
VQ-VAE$^\dagger$ & \textbf{0.002} & 25.273 & 0.172 & \textbf{0.004} & 15.466 & \underline{0.170} \\
GOS-VAE$^\dagger$ & \underline{0.005} & \textbf{17.517} & \textbf{0.066} & \underline{0.008} & \textbf{8.719} & \textbf{0.075} \\
\bottomrule
\end{tabular}
\vspace*{-1mm}
\label{tab:table3}
\end{table}

Next by comparing GOS-VAE$^{\dagger}$ to GOS-VAE$^{*}$, we see a notable improvement, particularly in mIoU. Additionally, GOS-VAE$^{\dagger}$ outperforms GESCO in mIoU and achieves comparable pixel accuracy. This experiment highlights that even a slight increase in computation could lead to substantial gains in downstream task results without extra bandwidth. Notably, when comparing the number of trainable parameters, the proposed GOS-VAE$^{\dagger}$ is about $92$ times smaller than GESCO.

To evaluate the generalizability of our GOS-VAE, we present the results of semantic segmentation on the ADE20K dataset, whose results are presented in Table \ref{tab:table2}. To ensure fair comparisons, we adjust the network structure to let each method maintain a similar bandwidth or payload. Our test results show that the proposed GOS-VAE$^{\dagger}$ substantially outperforms the existing schemes in 
each of the evaluation metrics, 
a result consistent with that from
test using Cityscapes.
These results further validate the effectiveness of the proposed GOS-VAEs when processing a larger and more complex dataset. Furthermore, the number of trainable parameters in this case is about $93$ times less than the diffusion-based GESCO.

Finally, in Semantic Communication, visual quality of reconstructed images is also essential; thus, we compare our proposed GOS-VAE$^{\dagger}$ with three representative methods. As shown in Table \ref{tab:table3}, GOS-VAE$^{\dagger}$ achieves superior visual quality and closer alignment to the original image distribution, demonstrated by lower FID and LPIPS scores, rather than prioritizing pixel-wise accuracy as measured by MSE.


\vspace{-1mm}
\subsection{Visualization Results}
\vspace{-1mm}
In addition to numerical metrics, Fig.~\ref{fig:visualization} presents visualization results for different approaches in image segmentation. From the results, we first notice that under limited transmission bandwidth, JPEG struggles to maintain good image quality and downstream task performance of the reconstructed images. Next, by comparing the proposed GOS-VAE$^{\dagger}$ to all other methods, we underscore the benefit and power of goal-oriented semantic communication framework over communication systems designed for bit-wise recovery. More specifically, the image reconstructed by GOS-VAE$^{\dagger}$ exhibits a high degree of visual consistency with the original one, even though it has a superficially higher pixel-wise Mean Squared Error (MSE) when compared against VQ-VAE. Moreover, in terms of the downstream segmentation task, GOS-VAE$^{\dagger}$ successfully detects small while important objects, such as traffic signs and traffic lights, in its predicted segmentation map. However, VQ-GAN, VQ-VAE, and GESCO are not able to fully detect these important objects. Furthermore, the boundaries of many objects in the segmentation maps are not accurate. These results further demonstrate that GOS-VAE$^{\dagger}$ is able to preserve the ``semantic information" defined by the downstream task.

\subsection{Ablation Study}
\vspace{-1mm}
\subsubsection{Performance with Different Objective Functions}
To illustrate the reasonability of our design of objective function for training our GOS-VAE as Eq. \eqref{eq:5}, we conduct an ablation study on different designs of loss function for GOS-VAE.

In the first setup, we use cross-entropy (CE) to measure the distribution similarity, denoted by GOS-VAE (CE), the loss can be characterized by

\vspace{-4mm}
\begin{equation}
L_{sc} = CE(S || \hat{S}) + \|\text{sg}[z_e(x)] - e\|_2^2 + \beta \|z_e(x) - \text{sg}[e]\|_2^2.
\end{equation}
\vspace{-4mm}

We also apply the Kullback–Leibler divergence (KLD) for distribution comparison, which can be applied to both original VQ-VAE and GOS-VAE. For example, the VQ-VAE (KLD) can be trained based upon 
\begin{align}
L_{vk} = & \, \frac{1}{N} \sum_{i=1}^{N} (x_i - \hat{x}_i)^2 + D_{KL}(S \parallel \hat{S}) \nonumber \\
        & + \|\text{sg}[z_e(x)] - e\|_2^2 + \beta \|z_e(x) - \text{sg}[e]\|_2^2,
\end{align}
while GOS-VAE (KLD + LPIPS) has a similar objective function as Eq. \eqref{eq:5} with the JSD replaced by the KLD.

For VQ-VAE (LPIPS), the pixel-wise Mean-Squared Error (MSE) is replaced with the LPIPS, while the downstream task loss is not included, calculated by
\begin{equation}
L_{vp} = \text{LPIPS}(x, \hat{x}) + \|\text{sg}[z_e(x)] - e\|_2^2 + \beta \|z_e(x) - \text{sg}[e]\|_2^2.
\end{equation}


\vspace{-2mm}
\begin{table}[htb]
\centering
\vspace{-2mm}
\caption{Ablation study on designing objective function using Cityscapes dataset.}
\vspace{-2mm}
\begin{tabular}{lcc}
\toprule
\textbf{Models}                          & \textbf{mIoU (\%) \textuparrow} & \textbf{Accuracy (\%) \textuparrow} \\
\midrule
GOS-VAE (CE)                           & 18.785                   & 81.715                      \\
GOS-VAE (KLD)                          & 42.792                   & 89.028                      \\
VQ-VAE (KLD)                    & 46.789                   & 89.759                      \\
VQ-VAE (LPIPS)                           & 54.933                   & 92.479                      \\
GOS-VAE (KLD + LPIPS)                     & {\underline{56.949}}          & {\underline{93.166}}             \\
GOS-VAE (JSD + LPIPS)                     & \textbf{57.342}          & \textbf{93.176}             \\
\bottomrule
\end{tabular}
\label{tab:table4}
\vspace*{-1mm}
\end{table}

As shown in Table \ref{tab:table4}, GOS-VAE (KLD+LPIPS) and GOS-VAE (JSD+LPIPS) consistently achieve the best performance, indicating the effectiveness of our proposed method on the downstream task. This can be attributed to the fact that the segmentation model, OneFormer, is pre-trained on the corresponding dataset, resulting in a data distribution that reflects images from the dataset. By utilizing LPIPS, data distribution of the reconstructed images aligns more closely with that of the OneFormer model, leading to improved performance. Compared to the CE-based method, KLD methods achieve better performance since KLD utilizes a distribution as the learning objective for each sample, providing more information than the one-hot label in Cross-Entropy.

\begin{figure}[htb]
\centering
\includegraphics[width=0.48\textwidth]{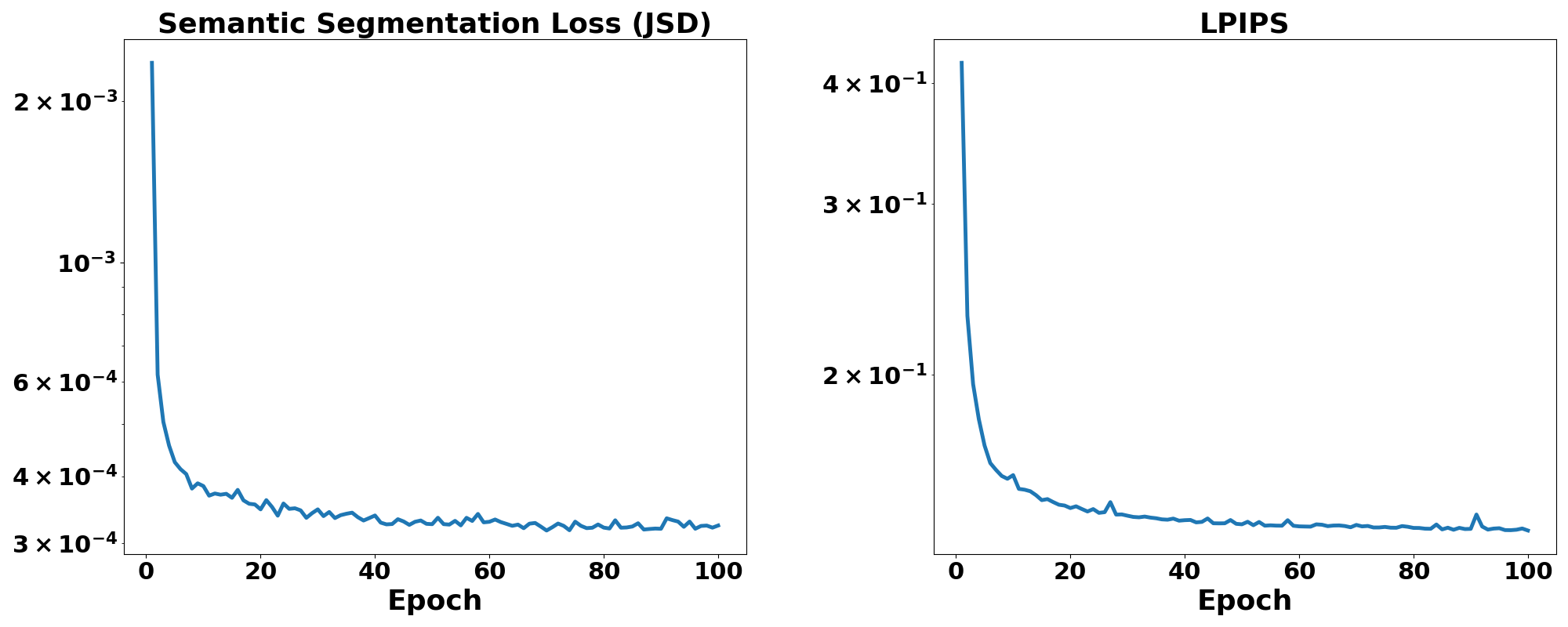}
\vspace*{-2mm}
\caption{Training curves of semantic segmentation loss (JSD) and LPIPS for GOS-VAE on the ADE20K dataset. The correlation of the two curves is 0.976.}
\label{fig:correlation_ade20k}
\vspace*{-1mm}
\end{figure}

We further analyze the relationship between semantic segmentation performance and LPIPS to further validate our conclusion. As shown in Fig. \ref{fig:correlation_ade20k}, the training curves for the two loss terms follow very similar trends with a correlation of $0.976$. This indicates that managing the data distribution shift is crucial for achieving optimal performance when using a pre-trained downstream task model to train an efficient compression network for GO-COM.

\begin{figure}[htb]
\centering
\vspace{-1mm}
\includegraphics[width=0.48\textwidth
]{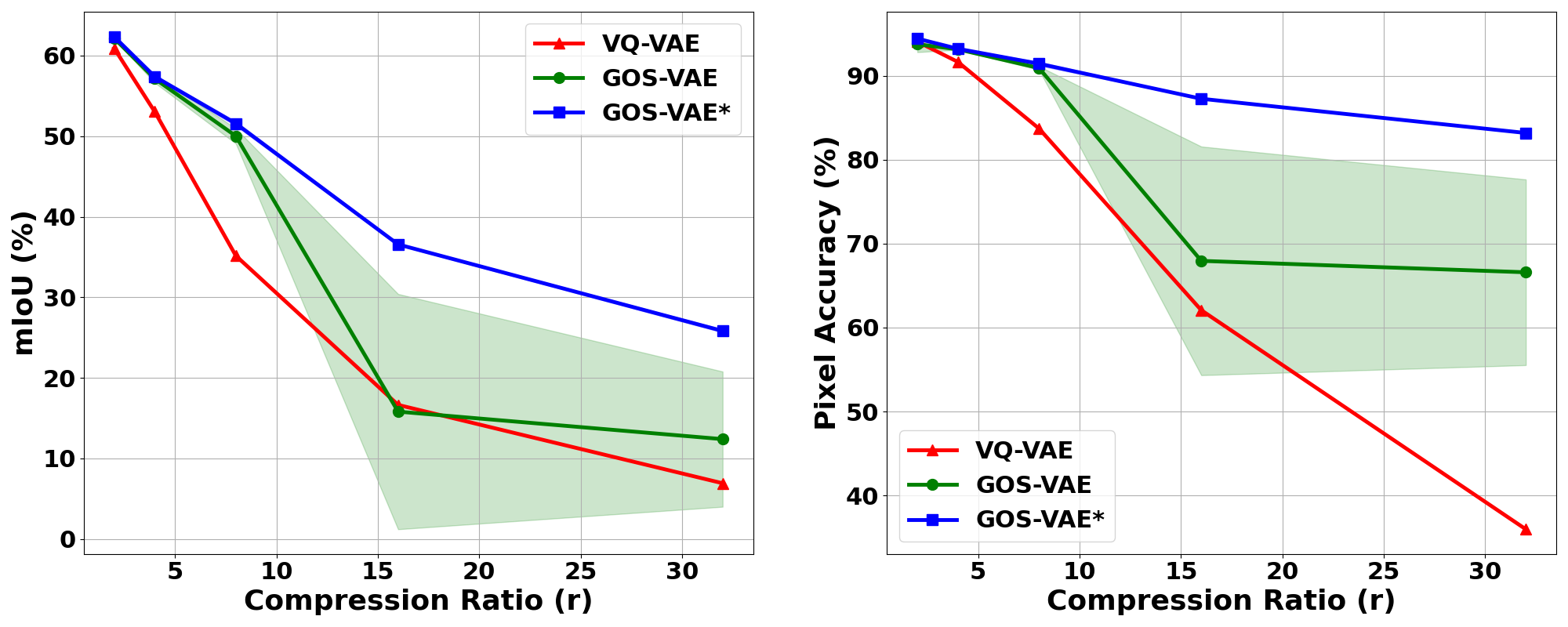}
\caption{Performance comparisons of models on the Cityscapes dataset under different compression ratios (r).}
\label{fig:compression_ratio}\vspace*{-2mm}
\end{figure}

\subsubsection{Performance for Different Compression Ratios}
The tradeoff between compression ratio and downstream task performance is of critical concern for efficient data transmission. In this test, we compare the performance of semantic segmentation of reconstructed images from different methods at various compression ratios (r):  $2$, $4$, $8$, $16$, and $32$. 
From the results presented in 
Fig.~\ref{fig:compression_ratio}, we observe that although GOS-VAE can achieve performance comparable to GOS-VAE$^{*}$, its stability 
suffers at higher compression ratios. Moreover, GOS-VAE may fail to converge in some cases, with the worst-case mIoU dropping to as low as $5.462$. The stability at higher compression ratios is the motivation for initializing GOS-VAE$^{*}$ with pre-trained VQ-VAE. On the other hand, GOS-VAE$^{*}$ consistently outperforms VQ-VAE across all compression ratios, establishing its
superior stability and robustness, particularly in terms of pixel accuracy.

\section{Conclusion}


In this work, we propose an innovative Goal-Oriented Semantic Variational Autoencoder (GOS-VAE) for task/goal-oriented communications, shifting the focus from visual quality or bit-wise recovery to conveying task-driven semantic features. Particularly, the proposed GOS-VAE defines ``semantic information" through end-to-end learning using a downstream task model and leverages imitation learning to preserve essential semantics in reconstructed images. Experimental results demonstrate that the proposed GOS-VAE framework delivers exceptional performance on the downstream task, surpassing previous methods while utilizing a low-complexity model architecture. Our future work plans to explore more advanced architectures for GO-COM, such as transformer and diffusion-based models, integrated with self-supervised learning.



\bibliographystyle{IEEEtran}
\bibliography{reference}

\end{document}